\RequirePackage{fix-cm}
\documentclass[a4paper,11pt]{article}

\usepackage{authblk}
\usepackage{fancyhdr}
\usepackage{amssymb}
\usepackage{microtype}
\usepackage{url}
\usepackage{tabularx}
\usepackage[table]{xcolor}
\usepackage{algorithm}
\usepackage{algorithmic}
\usepackage{multirow}
\usepackage{graphicx}
\usepackage{booktabs}
\usepackage{epsfig}
\usepackage{epstopdf}
\usepackage{amsmath}
\usepackage{amsfonts}
\usepackage{wasysym}
\usepackage{paralist}
\usepackage{url}
\usepackage{float}
\usepackage{paralist}
\usepackage{subcaption}
\usepackage{caption}
\usepackage{collcell}

\usepackage{adjustbox}

\providecommand{\keywords}[1]{\textbf{\textit{Keywords }} #1}

\begin{document}

\title{On the Genotype Compression and Expansion for Evolutionary Algorithms in the Continuous Domain}

\author[1]{Lucija Planinic}
\author[1]{Marko Djurasevic}
\author[2]{Luca Mariot}
\author[1]{Domagoj Jakobovic}
\author[2]{Stjepan Picek}
\author[3]{Carlos Coello Coello}

\affil[1]{{\normalsize Faculty of Electrical Engineering and Computing, University of Zagreb, Unska 3, Zagreb, Croatia} \\
	
	{\small \texttt{\{lucija.planinic,marko.djurasevic,domagoj.jakobovic\}@fer.hr}}}

\affil[2]{{\normalsize Cyber Security Research Group, Delft University of Technology, Mekelweg 2, Delft, The Netherlands} \\

  {\small \texttt{\{l.mariot, s.picek\}@tudelft.nl}}}

\affil[3]{{\normalsize CINVESTAV-IPN, Departamento de Computaci\'{o}n, Mexico City, Mexico} \\
	
	{\small \texttt{ccoello@cs.cinvestav.mx}}}

\maketitle

\begin{abstract}
  This paper investigates the influence of genotype size on evolutionary algorithms' performance. We consider genotype compression (where genotype is smaller than phenotype) and expansion (genotype is larger than phenotype) and define different strategies to reconstruct the original variables of the phenotype from both the compressed and expanded genotypes.
  We test our approach with several evolutionary algorithms over three sets of optimization problems: COCO benchmark functions, modeling of Physical Unclonable Functions, and neural network weight optimization. 
  Our results show that genotype expansion works significantly better than compression, and in many scenarios, outperforms the original genotype encoding.
  This could be attributed to the change in the genotype-phenotype mapping introduced with the expansion methods: this modification beneficially transforms the domain landscape and alleviates the search space traversal. 
\end{abstract}


\section{Introduction}
\label{sec:intro} 

The dimensionality of an optimization problem significantly impacts any optimization procedure used to solve it, including evolutionary algorithms (EAs)~\cite{Chen2015, Omidvar2011}. However, there is still an open question of whether the difficulties of EAs with high dimensional problems are mainly due to the larger number of variables or the problem becoming harder to solve. Often, these two factors are correlated: a higher number of variables increases the search space size, which in turn makes the problem more complex -- for example, by increasing the number of local minima, as in the case of rugged fitness landscapes~\cite{grundel07}. Therefore, from existing research, it is difficult to deduce how increasing the dimension space influences the algorithms' performance since changes in performance could also be due to an increase in difficulty. 
\thispagestyle{fancy}
This paper aims to observe how the performance of EAs correlates with different dimension sizes while keeping the same problem size. This means that the algorithms operate on a different number of variables than required by the optimization problem. These algorithm-side variables are then transformed in a specific way to obtain the correct number of problem-specific variables before the evaluation procedure. In this way, the complexity of the original problem is preserved (i.e., the problem does not become easier or more difficult to solve), whereas the search space size is artificially enlarged or reduced. Doing so makes it possible to obtain a more objective estimation of the influence of 
the problem dimensionality for the performance of optimization methods. 

Approaches that artificially increase or decrease the search space did not receive much attention in the literature yet. Salcedo-Sanz et al.~\cite{salcedo07} proposed a genotype compression-expansion strategy to improve the convergence speed of a Genetic Algorithm (GA) on the Inductive Query By Example (IQBE) optimization problem related to information retrieval systems. The GA genotype encodes the Boolean terms in a query, and the compression step groups in fixed-size subsets the terms belonging to the same topic, representing them with a single bit. The compressed individual is then expanded back for fitness evaluation. 
Koutnik et al.~\cite{Koutnik2010} considered the problem of reducing the search space in the context of neuroevolution. In that work, the weight matrix of a neural network is represented in the frequency domain utilizing a Fourier transform. Evolutionary algorithms are then used to search in this compressed representation, where high-frequency coefficients are removed. Steenkiste et al.~\cite{steenkiste16} adopted a similar approach but used a wavelet transform instead. In both cases, the underlying assumption for this compression strategy is that successful networks have spatially correlated weights. Moreno et al. ~\cite{moreno} proposed methods based on artificial neural network autoencoders for the automatic generation of genotype-phenotype mappings to improve evolvability. One of the proposed methods uses the encoder part of a bottlenecked autoencoder to reduce the genotype size and the decoder segment for the genotype-phenotype mapping. Stanley et al. ~\cite{Stanley2003ATF} developed a taxonomy for Artificial Embriogeny, a subdiscipline of evolutionary computation, where some genes from the genotype are used multiple times while mapping from genotype to phenotype. This allows for a simpler representation of a complex phenotype, which therefore reduces the search space. Similarly, Bongard et al. ~\cite{Bongard} used a developmental encoding scheme based on Artificial Ontogeny to map the genotype in a phenotype that represents a complete agent.  

From the above examples, the few works focusing on genotype compression and expansion mainly consider specific optimization problems, and the proposed strategies leverage domain-specific knowledge. As far as we know, this paper represents the first study on the influence of genotype compression/expansion for evolutionary algorithms, where the compression and expansion strategies are independent of the underlying optimization problem. Our experimental setup considers continuous optimization problems and several evolutionary algorithms to draw conclusions. More precisely, we consider three problems: the COCO benchmark functions, Physical Unclonable Functions (PUF) modeling, and neural network weights optimization. For the compression procedure, we investigate the scenario where two variables are combined into a single variable (reducing the dimensionality of the problem by half). The compression procedure is conducted via two strategies: interleaving and concatenation. We experiment with several sizes of the expanded variables for the expansion procedure and two procedures to obtain those: summation and multiplication.
Our results show that the compression procedure consistently gives poor results, but expansion can significantly improve the performance of the algorithm (as evaluated by the final solution).

\section{Genotype Compression and Expansion}
\label{sec:setup}

\subsection{Genotype Compression}

The compression strategies presume that a number of original variables are combined into a single \textit{compressed variable}. Let $m$ represent the number of original variables combined into a single one; then the total number of variables after the compression will be equal to $\lceil \frac{t}{m} \rceil$. 
Regardless of the original uncompressed variables' scope, we keep every value of the compressed variable in the interval $[0,1]$.
In the floating-point representation, the number of decimal places is denoted by $p$ and given by the precision of the number format used to store those values.
In our experiments, we used a standard double-precision format with 16 significant digits.
Since the values of the compressed variables are in $[0, 1]$ range, the decimal part of the compressed value is used to decode $m$ uncompressed numbers.
As a result, each original variable will use $d = \left \lceil \frac{p}{m} \right \rceil$ decimal digits of the compressed number.

A compressed variable is decompressed by the following two strategies: \textit{sequential} and \textit{alternating}. 
In the sequential strategy (concatenated), the decoding is performed so that the first $p/m$ digits represent the first original value, the following $p/m$ digits the next value, etc.

In the alternating scheme (interleaved), the digits of the compressed variable are distributed so that each $m$-th digit is used for recreating one original variable.
After the decomposition, in both schemes, the resulting values are additionally mapped to the desired domain interval with a simple linear transformation, i.e., the value $0$ represents the lower, while the value $1$ the upper bound.

Regardless of the decoding scheme, the described compression approach inevitably incurs a loss of precision to the original variables; in the above example, the original variable would have only half of the significant digits representing its value.
Depending on the actual problem domain, the loss of precision may affect the algorithm performance; unfortunately, this is not immediately evident in every case.
In all experiments in this paper, we limit the division factor to $m=2$, so the resulting precision is $p/2$ significant digits in the original domain.

\subsection{Genotype Expansion}

Let $t$ denote the number of variables of a particular continuous optimization problem, referred to as the \textit{original variables}.
In the genotype expansion, each original variable is represented with several \textit{expanded variables}; let $m$ represent the number of expanded variables for a single original variable.
The optimization algorithm operates on individuals comprised of expanded variables; only when evaluating a potential solution, the corresponding original values are recreated, and the fitness is calculated and assigned to the individual.

To recover the original variables, we consider the following two strategies: \textit{summation} and \textit{multiplication}. 
In the summation scheme, the original variable $x_i$ can be split into the sum of $m$ variables $x_i=x_{i1}+x_{i2}+\ldots+x_{im}$. In the multiplication scheme, the original variable $x_i$ is represented as a product of $m$ variables $x_i=x_{i1}\cdot x_{i2}\cdot \ldots \cdot x_{im}$. 
This means that the expanded representation can be decoded into the original variable by consecutively summing or multiplying $m$ expanded variables. In both cases, $m$ can be an arbitrary integer number larger than 1. The number of variables in the expanded scheme is then $t\cdot m$.

The expansion strategies were selected with both simplicity and efficiency in mind to demonstrate the applicability of such an approach, although more complex transformations can be defined. However, one potential pitfall of the multiplication strategy has to be outlined. When the allowed domain is defined only between 0 and 1, then as the number of multiplications increases, the result will approach 0. This can be resolved either by increasing the domain to have multiplications with larger numbers or by decreasing the number of multiplications.

The expanded variables use the same domain (defined with lower and upper bound) as the original variables.
The decoded original value can be clipped to the bound value if it exceeds a specific interval to facilitate constrained optimization with explicit bounds. Unlike in ~\cite{Koutnik2010} where certain elements are discarded from the search space, both proposed strategies use the entire genotype during the transformation to phenotype. In that way, all elements are directly considered during optimization.  

\section{Experimental Results and Discussion}
\label{sec:results}

\subsection{Experimental Setup}

The genotype expansion or compression is used independently of the optimization algorithm; however, the efficiency of the approach may still be influenced by the chosen algorithm.
We have performed the experiments using several well-known evolutionary algorithms, each using the original variables (denoted as ``default''), the expanded, and the compressed variables.
We investigate the genetic algorithm (GA), evolutionary strategy (ES), differential evolution (DE), and clonal selection algorithm (CLONALG). However, as GA and DE achieved the best results and the tested approaches exhibited a similar behavior on the other algorithms as well, we report the detailed results only for GA and DE.

The GA uses a steady-state selection scheme where, in each iteration, three individuals are selected at random. The worst one is eliminated and replaced with the crossover offspring from the remaining two.
The resulting new individual is additionally mutated with a mutation rate of 0.3.
The differential evolution uses a scaling constant (differential weight) $F = 1$ and the crossover rate $CR = 0.9$.
The population size in all experiments is set to 100.
The parameters we use are selected based on a short preliminary tuning.

All the algorithms operate on the same genotype, where individuals are represented as vectors of floating-point values.
In GA, a single mutation operator is used, which alters a randomly chosen gene (a single floating-point value) to a new random value from the domain with a uniform distribution.
For the crossover in GA, we use multiple operators where a random one is selected every time crossover is performed.
The crossover operator is selected from the following: discrete crossover~\cite{Eiben03}, simple and whole arithmetic crossover~\cite{Eiben03}, local crossover~\cite{Dumitrescu00}, SBX crossover~\cite{boyer}\cite{agrawal}, BLX-alpha crossover~\cite{Eshelman92}, flat crossover~\cite{radcliffe}, BGA crossover~\cite{muhlenbein}, heuristic crossover\cite{wright}, and average crossover~\cite{nomura}.

We use the Mann-Whitney non-parametric test to check whether the proposed approaches' results are significantly better than the standard (default) number of variables (pair-wise comparison). First, we test the hypothesis that the proposed approach is significantly better than the standard one. If this hypothesis is rejected, we perform an additional test to check whether the proposed approach is significantly worse than the default one. The results are considered significant if the obtained $p$-values are less than 0.05. Each table will include the results of the statistical tests between the proposed approach and the default approach. The test results are listed after each value in the tables and denoted as +, - or =, representing the result is significantly better, significantly worse, or that there is no significant difference, respectively. 

\subsection{Benchmark Problems}
The COCO platform is used to analyze the performance of the proposed expansion/compression approaches~\cite{hansen2016}. From this platform, the set of 24 noiseless functions of arbitrary variable sizes is selected for the experiments. Based on their properties, the functions can be grouped into five categories:
\begin{compactitem}
	\item Separable functions - variables of the function are mutually independent, meaning that the optimum can be obtained by performing optimization in each dimension separately (functions $f1$ - $f5$).
	\item Functions with low or moderate conditioning - non-separable unimodal functions (which contain a single minimum), where a small change in the input variables does not lead to a large change in the function value (functions $f6$ - $f9$).
	\item Unimodal functions with high conditioning - non-separable unimodal functions, where a small change in the input variables leads to a large change in the function value, making it more difficult to obtain the correct solution (functions $f10$ - $f14$).
	\item Multi-modal functions with adequate global structure - non-separable functions with multiple minima that are uniformly distributed over the search space (functions $f15$ - $f19$).
	\item Multi-modal functions with weak global structure - non-separable functions with multiple minima with a non-uniform distribution over the search space (functions $f20$ - $f24$).
\end{compactitem}

The benchmark problems were optimized with all four algorithms, but the results are only shown for GA and DE.
Each algorithm used the maximum number of function evaluations as the stopping criterion, which is set to the value of $D \cdot 100\,000$, where $D$ represents the number of dimensions for the considered problems. The approaches are benchmarked on 2, 5, 10, and 20-dimensional functions. The search space is defined as the interval $[-5,5]$ in every dimension. For each tested function, 50 instances are created in the standard procedure used by the COCO platform by shifting or rotating the functions (for example, the separable functions are only shifted). Additionally, the COCO platform defines a run as successful if an objective value of $1e-8$ or less is reached, which is then denoted as a hit.

The experiments indicate that the GA obtained the best results among the tested methods, followed by DE. The GA results will be outlined for most of this paper, as the observations made there are also applicable for the other tested algorithms. 

Table~\ref{tab:coco_2} shows the results obtained by GA for optimizing the set of test functions in two dimensions. 
The columns in the table represent different encodings, with ``\textit{exp}'' corresponding to the expansion and ``\textit{com}'' to genotype compression, while ``\textit{def}'' denotes the default encoding in which the number of variables is unchanged.
We also report the expansion factor (2 or 3) and the decoding scheme; summation (\textit{s}) and multiplication (\textit{m}) for the expansion, sequential (\textit{seq}), and alternating (\textit{alt}) for the compression.
Each row denotes the median fitness value obtained over the 50 executed instances, the number of hits for each function (denoted in brackets), and whether the approach in this column is significantly better, equal, or worse concerning the default genotype encoding. The experiments in which the proposed methods obtained significantly better results than the default encoding are emphasized with grey cells.

It can be observed that compressing the search space to fewer variables leads to poor results, even for the simplest functions. This is backed up by statistical tests, demonstrating that the compression-based approaches achieve significantly worse results than the default encoding with the original number of variables for each function. 
For the expansion of variables, the results are more interesting. For the group of separable functions, the results show that all variants achieved the maximum number of hits. This shows that none of the expansion approaches has difficulties converging to the minimum on simple functions. On the second group of functions, we see that the expanded representations achieve a larger number of hits than the default method, except for function f7, where all methods achieved the maximum number of hits. Additionally, the expansion-based approaches obtained significantly better results in most cases for the other functions. As such, the proposed approaches perform better on non-separable functions that are not ill-conditioned. 

In the next group of functions, we observe a difference between the effectiveness of the summation and multiplication-based extension. While the summation-based approach has a similar performance to the default one, the multiplication-based one encounters problems and has a smaller number of hits, and in some cases, obtains significantly worse results. For unimodal functions (functions f1-f5), there is little difference between the default and summation-based approach, as both can obtain the optimum. 
For the third group of functions, we observe no significant difference between the summation-based and default approaches. We postulate that this happens as those functions exhibit a strong dependence between small changes in the specific gene values and the fitness value.
On the fourth group of functions, multi-modal with an adequate global structure, the expansion-based approaches achieve their best results compared to the default one. The summation-based approach obtained significantly better results for all five functions, whereas the multiplication-based approach is slightly less successful. On the final set of functions (weak global structure), the improvements over the default approach are less prominent but present in two cases.
To conclude, the proposed expansion approaches seem to perform better for multi-modal problems than the default approach, while for unimodal cases, the differences are smaller.

\begin{table*}[]
	\small
	\caption{Results obtained by GA for the COCO benchmark with 2 dimensional functions.}
	\label{tab:coco_2}
	\adjustbox{max width=\textwidth}{
		\begin{tabular}{@{}clllllll@{}}
			\toprule
			\multicolumn{1}{l}{} & \multicolumn{1}{c}{\textbf{def}} & \multicolumn{1}{c}{\textbf{exp-s-2}}                           & \multicolumn{1}{c}{\textbf{exp-s-3}}                        & \multicolumn{1}{c}{\textbf{exp-m-2}}                              & \multicolumn{1}{c}{\textbf{exp-m-3}}                           & \multicolumn{1}{c}{\textbf{com-seq}} & \multicolumn{1}{c}{\textbf{com-alt}} \\ \midrule
			\textbf{f1}          & 0.00e+00 (50)                      & 0.00e+00 (50) =                                              & 0.00e+00 (50) =                                              & 0.00e+00 (50) +                       & 0.00e+00 (50) +                       & 5.51e-05 (0) -                     & 7.99e-08 (22) -                    \\
			\textbf{f2}          & 0.00e+00 (50)                      & 0.00e+00 (50) =                                              & 0.00e+00 (50) -                                              & 0.00e+00 (50) =                                              & 0.00e+00 (50) =                                              & 1.57e-02 (0) -                     & 1.87e-01 (0) -                     \\
			\textbf{f3}          & 0.00e+00 (50)                      & 0.00e+00 (50) =                                              & 0.00e+00 (50) =                                              & 0.00e+00 (50) +                       & 0.00e+00 (50) =                                              & 4.88e-02 (0) -                     & 8.27e-04 (1) -                     \\
			\textbf{f4}          & 0.00e+00 (50)                      &  0.00e+00 (50) +                    &  0.00e+00 (50) +                  &  0.00e+00 (50) +                      & 0.00e+00 (50) +                                             & 1.05e-01 (0) -                     & 2.29e-03 (1) -                     \\
			\textbf{f5}          & 0.00e+00 (50)                      & 0.00e+00 (50) =                                              & 0.00e+00 (50) =                                              & 0.00e+00 (50) =                                              & 0.00e+00 (50) =                                              & 6.93e-03 (23) -                    & 1.20e-06 (23) -                    \\ \midrule
			\textbf{f6}          & 3.37e-09 (31)                      & \cellcolor[HTML]{dfdfdf}2.38e-10 (43) + & \cellcolor[HTML]{dfdfdf}4.88e-12 (50) + & \cellcolor[HTML]{dfdfdf}3.03e-10 (45) + & \cellcolor[HTML]{dfdfdf}1.10e-10 (48) + & 2.79e-03 (0) -                     & 2.81e-04 (1) -                     \\
			\textbf{f7}          & 0.00e+00 (50)                      & 0.00e+00 (50) =                                              & 0.00e+00 (50) =                                              & 0.00e+00 (50) =                                              & 0.00e+00 (50) =                                              & 3.19e-08 (12) -                    & 2.24e-13 (48) -                    \\
			\textbf{f8}          & 2.73e-09 (31)                      & \cellcolor[HTML]{dfdfdf}1.98e-11 (44) + & \cellcolor[HTML]{dfdfdf}3.84e-12 (42) + & \cellcolor[HTML]{dfdfdf}6.14e-11 (40) + & 2.71e-09 (34) =                                              & 5.81e-04 (0) -                     & 9.48e-05 (1) -                     \\ 
			\textbf{f9}          & 1.84e-10 (34)                      & 1.55e-11 (43) =                                              & \cellcolor[HTML]{dfdfdf}1.46e-12 (46) + & \cellcolor[HTML]{dfdfdf}2.86e-11 (45) + & \cellcolor[HTML]{dfdfdf}2.43e-12 (46) + & 8.22e-04 (0) -                     & 9.44e-05 (3) -                     \\ \midrule
			\textbf{f10}         & 9.82e-06 (9)                       & 3.67e-06 (9) =                                               & 2.83e-05 (6) =                                               & 5.00e-04 (5) -                                               & 5.11e-03 (2) -                                               & 1.75e-01 (0) -                     & 8.49e-02 (0) -                     \\
			\textbf{f11}         & 2.69e-06 (10)                      & 9.29e-07 (14) =                                              & 1.60e-06 (13) =                                              & 4.81e-03 (10) -                                              & 4.62e-04 (8) -                                               & 1.18e-01 (0) -                     & 9.70e-02 (0) -                     \\
			\textbf{f12}         & 3.48e-05 (15)                      & 6.92e-06 (20) =                                              & 3.07e-05 (19) =                                              & 1.28e-04 (19) =                                              & 1.52e-04 (17) =                                              & 1.97e-01 (0) -                     & 8.26e-02 (0) -                     \\
			\textbf{f13}         & 6.11e-05 (0)                       & 3.74e-05 (2) =                                               & 5.86e-05 (0) =                                               & 2.98e-04 (2) -                                               & 1.48e-04 (0) =                                               & 7.19e-02 (0) -                     & 3.79e-02 (0) -                     \\
			\textbf{f14}         & 7.32e-07 (7)                       & 4.75e-07 (10) =                                              & 9.27e-07 (7) =                                               & 1.78e-06 (8) =                                               & 1.80e-06 (3) =                                               & 7.30e-04 (0) -                     & 8.21e-05 (0) -                     \\ \midrule
			\textbf{f15}         & 8.11e-11 (39)                      & \cellcolor[HTML]{dfdfdf}0.00e+00 (48) + & \cellcolor[HTML]{dfdfdf}0.00e+00 (50) + & \cellcolor[HTML]{dfdfdf}0.00e+00 (48) + & \cellcolor[HTML]{dfdfdf}0.00e+00 (47) + & 2.93e-02 (0) -                     & 3.15e-04 (4) -                     \\
			\textbf{f16}         & 1.32e-11 (32)                      & \cellcolor[HTML]{dfdfdf}0.00e+00 (36) + & \cellcolor[HTML]{dfdfdf}0.00e+00 (41) + & 0.00e+00 (36) =                                              & \cellcolor[HTML]{dfdfdf}0.00e+00 (37) + & 4.43e-03 (0) -                     & 4.29e-04 (3) -                     \\
			\textbf{f17}         & 3.32e-05 (13)                      & \cellcolor[HTML]{dfdfdf}2.04e-10 (29) + & \cellcolor[HTML]{dfdfdf}2.18e-11 (31) + & \cellcolor[HTML]{dfdfdf}1.21e-12 (30) + & \cellcolor[HTML]{dfdfdf}1.60e-14 (32) + & 1.94e-02 (0) -                     & 3.64e-03 (0) -                     \\
			\textbf{f18}         & 9.90e-04 (0)                       & \cellcolor[HTML]{dfdfdf}9.85e-04 (2) +  & \cellcolor[HTML]{dfdfdf}1.62e-04 (6) +  & 9.87e-04 (1) =                                               & 9.90e-04 (1) =                                               & 9.78e-02 (0) -                     & 3.25e-02 (0) -                     \\
			\textbf{f19}         & 1.93e-12 (50)                      & \cellcolor[HTML]{dfdfdf}2.66e-14 (50) + & \cellcolor[HTML]{dfdfdf}5.95e-14 (50) + & \cellcolor[HTML]{dfdfdf}3.55e-15 (49) + & \cellcolor[HTML]{dfdfdf}7.11e-15 (50) + & 9.43e-06 (0) -                     & 7.56e-07 (7) -                     \\ \midrule
			\textbf{f20}         & 2.23e-12 (50)                      & \cellcolor[HTML]{dfdfdf}0.00e+00 (50) + & \cellcolor[HTML]{dfdfdf}0.00e+00 (50) + & \cellcolor[HTML]{dfdfdf}0.00e+00 (50) + & \cellcolor[HTML]{dfdfdf}0.00e+00 (50) + & 3.64e-03 (0) -                     & 9.84e-04 (3) -                     \\
			\textbf{f21}         & 0.00e+00 (50)                      & 0.00e+00 (50) =                                              & 0.00e+00 (50) =                                              & 0.00e+00 (50) =                                              & 0.00e+00 (50) =                                              & 2.34e-08 (18) -                    & 2.04e-12 (45) -                    \\
			\textbf{f22}         & 0.00e+00 (50)                      & 0.00e+00 (50) =                                              & 0.00e+00 (50) =                                              & 0.00e+00 (50) =                                              & 0.00e+00 (50) =                                              & 5.79e-08 (14) -                    & 5.75e-10 (35) -                    \\
			\textbf{f23}         & 4.86e-03 (6)                       & \cellcolor[HTML]{dfdfdf}7.70e-04 (17) + & \cellcolor[HTML]{dfdfdf}2.30e-04 (16) + & \cellcolor[HTML]{dfdfdf}1.52e-04 (15) + & \cellcolor[HTML]{dfdfdf}5.35e-06 (21) + & 1.90e-01 (0) -                     & 2.02e-01 (0) -                     \\
			\textbf{f24}         & 6.09e-02 (10)                      & 5.84e-02 (11) =                                              & 5.92e-02 (14) =                                              & 2.81e-02 (16) =                                              & 2.37e-02 (11) =                                              & 3.46e-01 (0) -                     & 1.57e-01 (0) -                     \\ \bottomrule
	\end{tabular}}
\end{table*}

Unfortunately, as the number of dimensions increases, the difference between the results becomes less prominent; this can be observed from Table~\ref{tab:coco_20}, which shows the results obtained when optimizing 20-dimensional functions.
The compression-based approaches still achieve inferior results, whereas the expansion-based approaches, in most cases, obtain results that are not significantly different from the reference encoding. In this case, the expansion approach based on summation achieves equally good results as without expansion. However, this is not the case with the expansion based on multiplication, which achieves significantly worse results for around 40\% of tested functions, both when the number of variables is doubled and tripled. It can be concluded that the summation-based approach scales better with the dimensionality of the problem. Still, note that relatively poor results here do not necessarily mean that the expansion procedure does not work. The default approach results suggest that the considered optimization procedures are not powerful enough for high-dimensional problems (at least with the experimental setup as considered in this paper).

\begin{table*}[]
	\small
	\caption{Results obtained by GA for the COCO benchmark with 20 dimensional functions.}
	\label{tab:coco_20}
	\adjustbox{max width=\textwidth}{
		\begin{tabular}{@{}clllllll@{}}
			\toprule
			\multicolumn{1}{l}{} & \multicolumn{1}{c}{\textbf{def}} & \multicolumn{1}{c}{\textbf{exp-s-2}}                           & \multicolumn{1}{c}{\textbf{exp-s-3}}                        & \multicolumn{1}{c}{\textbf{exp-m-2}}                              & \multicolumn{1}{c}{\textbf{exp-m-3}}                           & \multicolumn{1}{c}{\textbf{com-seq}} & \multicolumn{1}{c}{\textbf{com-alt}} \\ \midrule
			\textbf{f1} & 3.98e-12 (50) & 6.85e-12 (50) = & 6.43e-12 (50) = & 6.27e-12 (50) = & 5.68e-12 (50) = & 9.28e-04 (0) - & 5.58e-05 (0) - \\
			\textbf{f2} & 3.79e-09 (38) & 2.49e-09 (34) = & 4.72e-09 (37) = & 2.50e-09 (35) = & \cellcolor[HTML]{dfdfdf}1.24e-09 (42) + & 9.84e+01 (0) - & 3.42e+01 (0) - \\
			\textbf{f3} & 5.94e-09 (34) & 6.28e-09 (33) = & 6.19e-09 (30) = & 1.23e-08 (25) - & 8.85e-09 (29) = & 9.56e-01 (0) - & 1.90e-01 (0) - \\
			\textbf{f4} & 6.12e-08 (3) & 5.89e-08 (6) = & 5.83e-08 (6) = & 5.81e-08 (8) = & 5.92e-08 (5) = & 2.35e+00 (0) - & 1.55e+00 (0) - \\
			\textbf{f5} & 9.95e-14 (50) & \cellcolor[HTML]{dfdfdf}0.00e+00 (50) + & \cellcolor[HTML]{dfdfdf}0.00e+00 (50) + & \cellcolor[HTML]{dfdfdf}0.00e+00 (50) + & \cellcolor[HTML]{dfdfdf}0.00e+00 (50) + & 5.67e-01 (0) - & 8.77e-03 (0) - \\ \midrule
			\textbf{f6} & 2.50e-02 (0) & 2.50e-02 (0) = & 2.46e-02 (0) = & 3.97e-01 (0) - & 2.51e-01 (0) - & 1.57e+00 (0) - & 1.58e+00 (0) - \\
			\textbf{f7} & 9.97e+00 (0) & 9.94e+00 (0) = & 9.96e+00 (0) = & 1.50e+01 (0) - & 1.58e+01 (0) - & 1.55e+01 (0) = & 1.25e+01 (0) = \\
			\textbf{f8} & 6.31e+00 (0) & 9.98e+00 (0) = & 1.58e+01 (0) = & \cellcolor[HTML]{dfdfdf}1.75e+00 (0) + & 3.25e+00 (0) = & 2.35e+01 (0) - & 2.40e+01 (0) - \\
			\textbf{f9} & 1.58e+01 (0) & 1.58e+01 (0) = & 1.58e+01 (0) = & \cellcolor[HTML]{dfdfdf}1.58e+01 (0) + & \cellcolor[HTML]{dfdfdf}1.58e+01 (0) + & 2.47e+01 (0) - & 2.48e+01 (0) - \\ \midrule
			\textbf{f10} & 2.51e+03 (0) & 2.51e+03 (0) = & 2.51e+03 (0) = & 3.98e+03 (0) - & 3.98e+03 (0) - & 1.58e+04 (0) - & 1.58e+04 (0) - \\
			\textbf{f11} & 3.98e+00 (0) & 3.97e+00 (0) = & 5.65e+00 (0) = & 1.58e+01 (0) - & 1.57e+01 (0) - & 1.51e+01 (0) - & 9.98e+00 (0) - \\
			\textbf{f12} & 9.63e-01 (0) & 2.50e+00 (0) - & 1.58e+00 (0) = & 2.51e+00 (0) - & 2.48e+00 (0) - & 9.38e+02 (0) - & 9.78e+01 (0) - \\
			\textbf{f13} & 2.51e+00 (0) & 3.97e+00 (0) = & 2.51e+00 (0) = & 2.50e+00 (0) = & 3.97e+00 (0) = & 1.52e+01 (0) - & 9.89e+00 (0) - \\
			\textbf{f14} & 1.00e-03 (0) & 1.00e-03 (0) = & 1.00e-03 (0) = & 1.00e-03 (0) = & 1.57e-03 (0) = & 2.43e-02 (0) - & 1.57e-02 (0) - \\ \midrule
			\textbf{f15} & 6.30e+01 (0) & 6.29e+01 (0) = & \cellcolor[HTML]{dfdfdf}6.20e+01 (0) + & 9.96e+01 (0) - & 1.57e+02 (0) - & 9.87e+01 (0) - & 6.31e+01 (0) - \\
			\textbf{f16} & 7.54e+00 (0) & 7.98e+00 (0) = & 9.43e+00 (0) = & 6.29e+00 (0) = & 7.87e+00 (0) = & 9.41e+00 (0) = & 9.94e+00 (0) - \\
			\textbf{f17} & 1.56e+00 (0) & 1.55e+00 (0) = & 1.58e+00 (0) = & 2.51e+00 (0) - & 3.96e+00 (0) - & 1.58e+00 (0) - & 1.58e+00 (0) - \\
			\textbf{f18} & 3.98e+00 (0) & 3.98e+00 (0) = & 3.97e+00 (0) = & 9.93e+00 (0) - & 1.50e+01 (0) - & 6.13e+00 (0) = & 6.10e+00 (0) = \\ 
			\textbf{f19} & 1.00e+00 (0) & 1.58e+00 (0) = & 1.58e+00 (0) - & \cellcolor[HTML]{dfdfdf}6.30e-01 (0) + & \cellcolor[HTML]{dfdfdf}2.51e-01 (0) + & 3.94e+00 (0) - & 3.89e+00 (0) - \\ \midrule
			\textbf{f20} & 5.57e-01 (0) & 4.71e-01 (0) = & 3.97e-01 (0) = & 9.67e-01 (0) - & 9.82e-01 (0) - & 3.98e-01 (0) = & 6.21e-01 (0) - \\
			\textbf{f21} & 2.49e+00 (15) & 2.45e+00 (16) = & 2.51e+00 (10) = & 2.50e+00 (15) = & 2.51e+00 (11) = & 2.39e+00 (0) = & 2.51e+00 (0) = \\
			\textbf{f22} & 9.73e+00 (0) & 6.18e+00 (0) = & 6.23e+00 (0) = & 9.64e+00 (0) = & 6.29e+00 (0) = & \cellcolor[HTML]{dfdfdf}3.98e+00 (0) + & 3.90e+00 (0) = \\
			\textbf{f23} & 1.51e+00 (0) & 1.00e+00 (0) = & 1.00e+00 (0) = & 9.99e-01 (0) = & \cellcolor[HTML]{dfdfdf}9.93e-01 (0) + & 1.43e+00 (0) = & \cellcolor[HTML]{dfdfdf}9.99e-01 (0) + \\
			\textbf{f24} & 6.29e+01 (0) & 6.30e+01 (0) - & 6.30e+01 (0) = & 9.99e+01 (0) - & 1.00e+02 (0) - & 1.20e+02 (0) - & 9.96e+01 (0) - \\ \bottomrule
	\end{tabular}}
\end{table*}

\begin{figure*}
	\centering
	\begin{subfigure}[b]{0.48\textwidth}
		\centering
		\includegraphics[width=\textwidth]{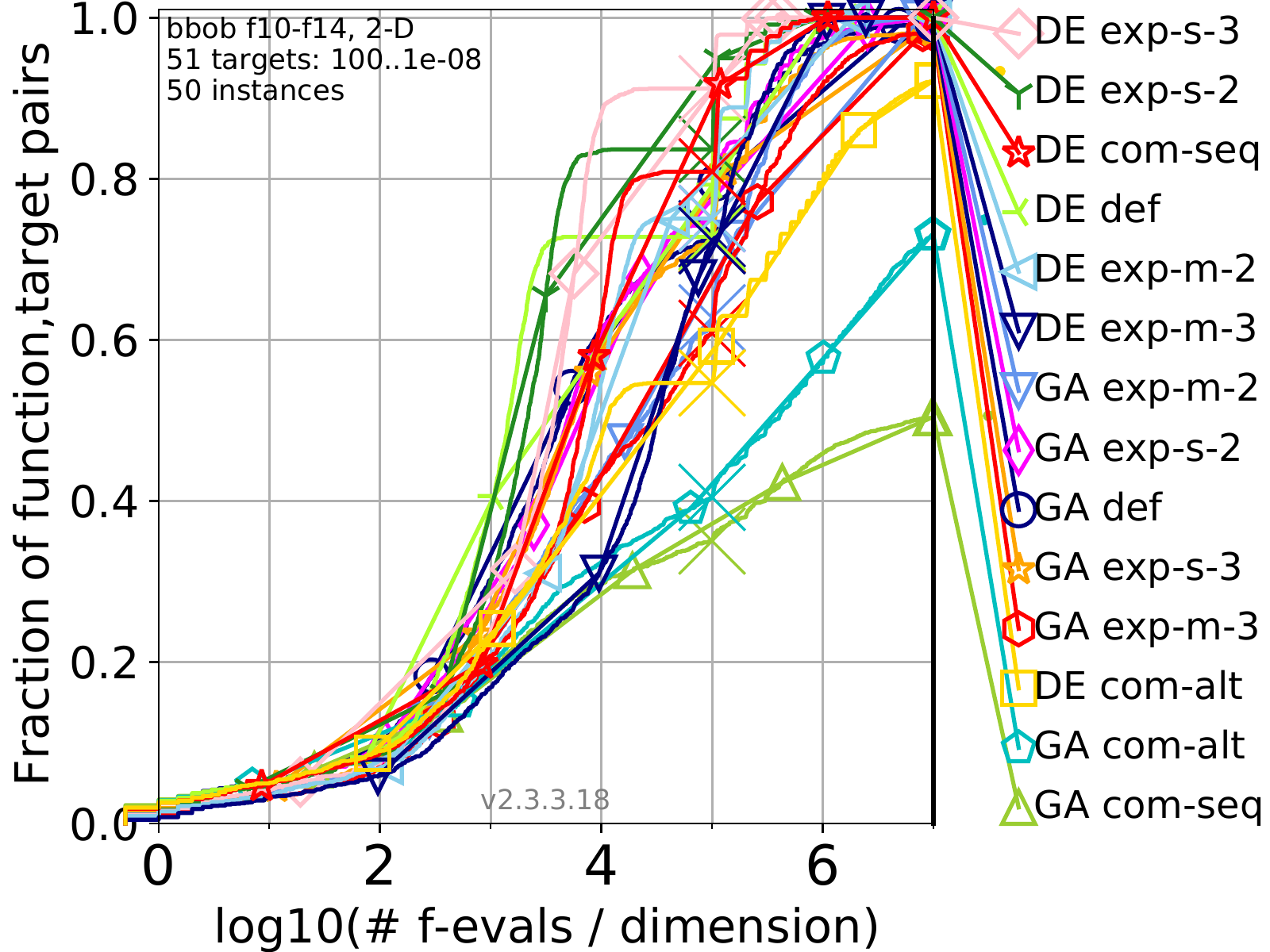}
		\caption{all functions}
		\label{fig:all}
	\end{subfigure}
	\begin{subfigure}[b]{0.48\textwidth}
		\centering
		\includegraphics[width=\textwidth]{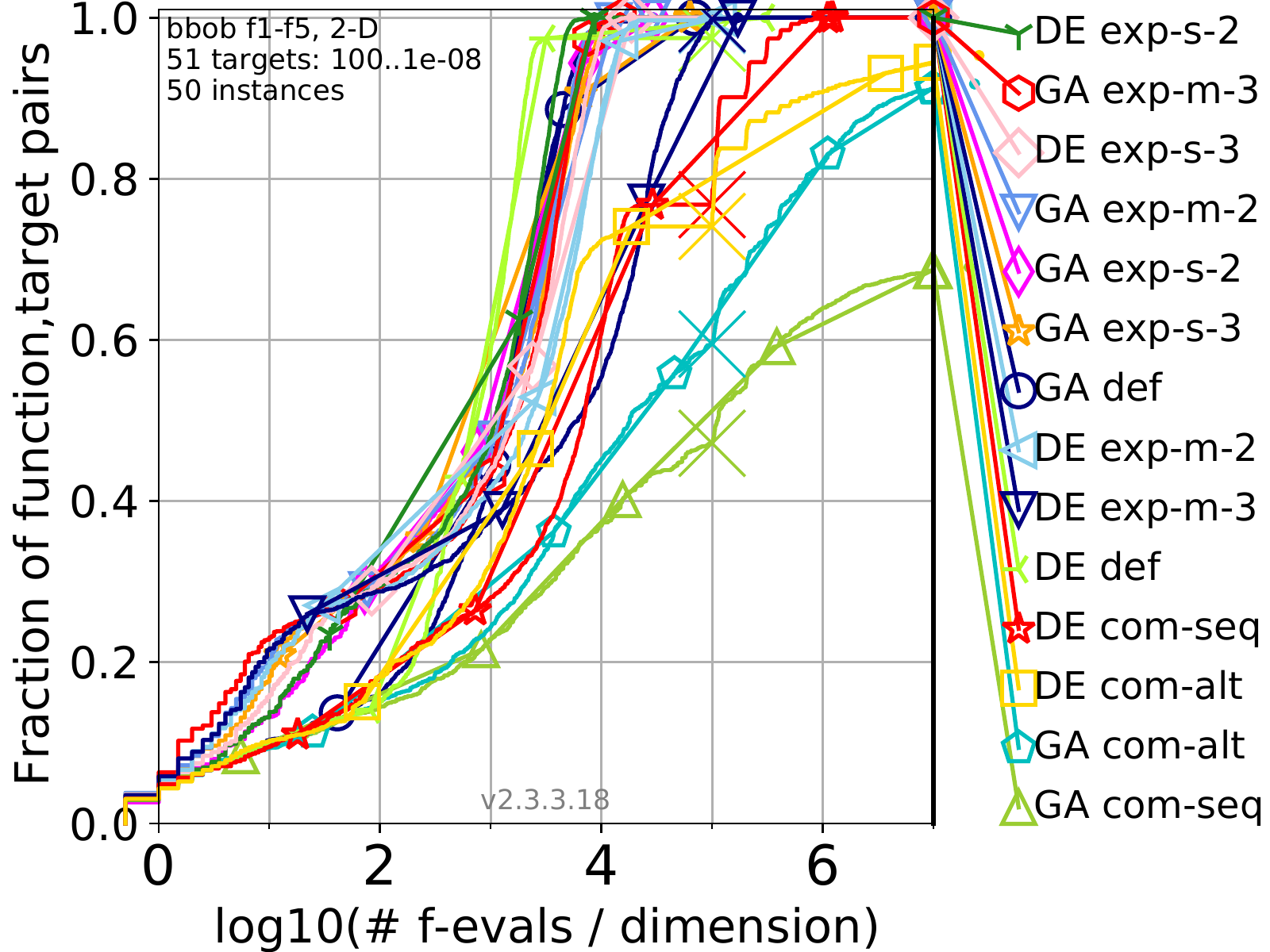}
		\caption{separable functions}
		\label{fig:separable}
	\end{subfigure}
	\begin{subfigure}[b]{0.48\textwidth}
		\centering
		\includegraphics[width=\textwidth]{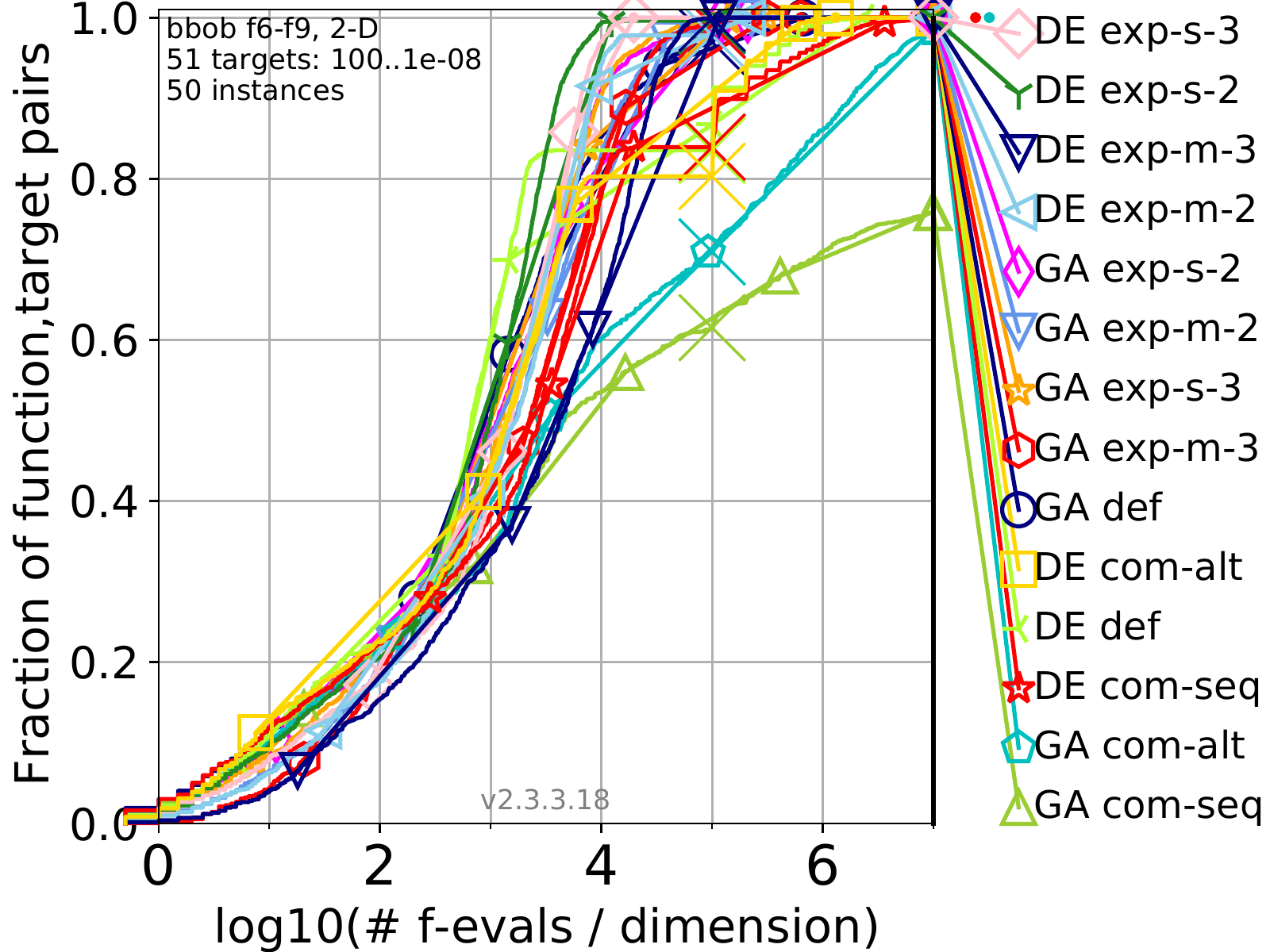}
		\caption{functions with low or moderate conditioning}
		\label{fig:low conditioning}
	\end{subfigure}
	\begin{subfigure}[b]{0.48\textwidth}
		\centering
		\includegraphics[width=\textwidth]{figures/pprldmany_02D_hcond.pdf}
		\caption{functions with high conditioning and unimodal}
		\label{fig:high conditioning}
	\end{subfigure}
	\begin{subfigure}[b]{0.48\textwidth}
		\centering
		\includegraphics[width=\textwidth]{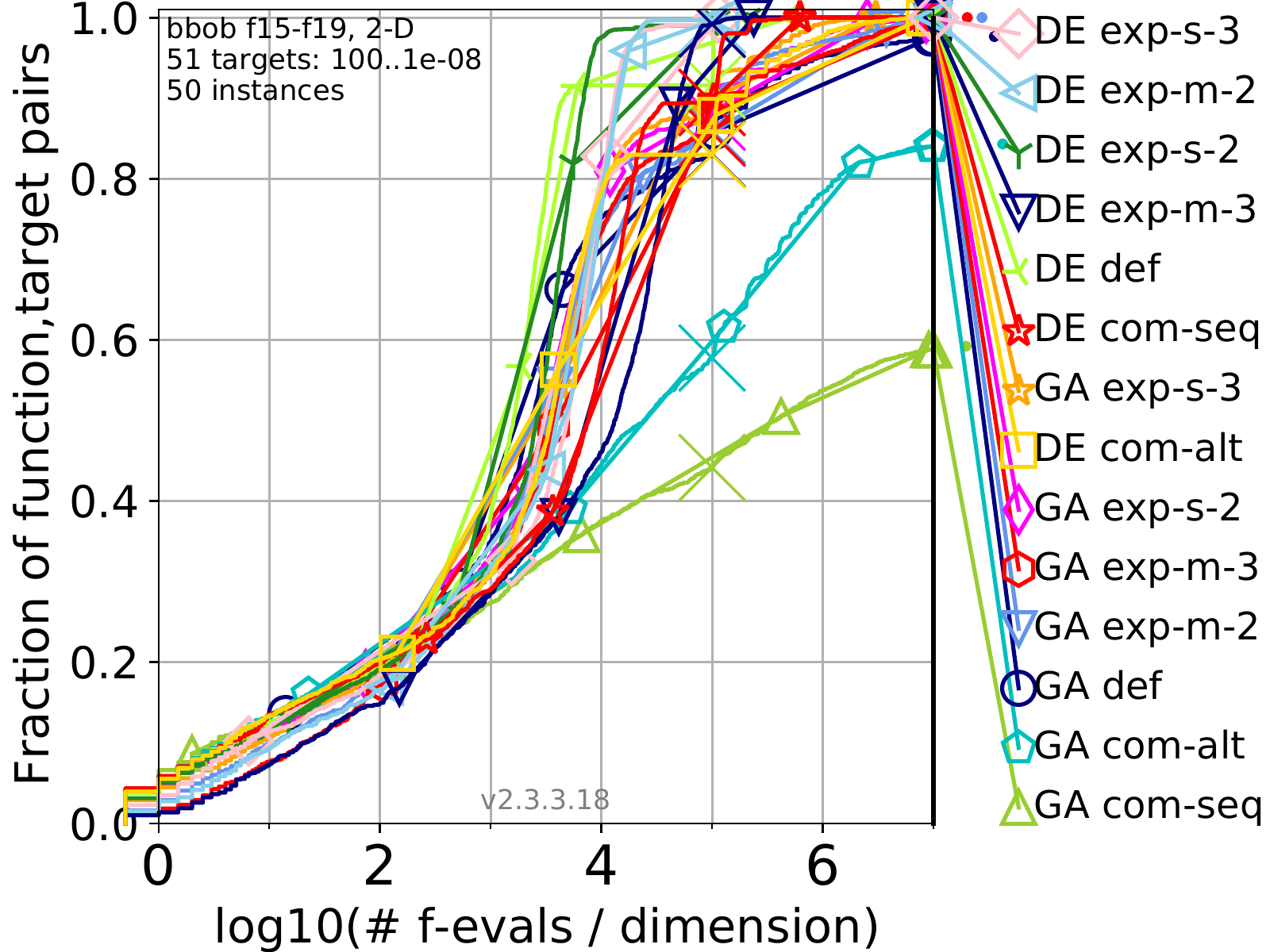}
		\caption{multi-modal functions with adequate global structure}
		\label{fig:adequate global structure}
	\end{subfigure}
	\begin{subfigure}[b]{0.48\textwidth}
		\centering
		\includegraphics[width=\textwidth]{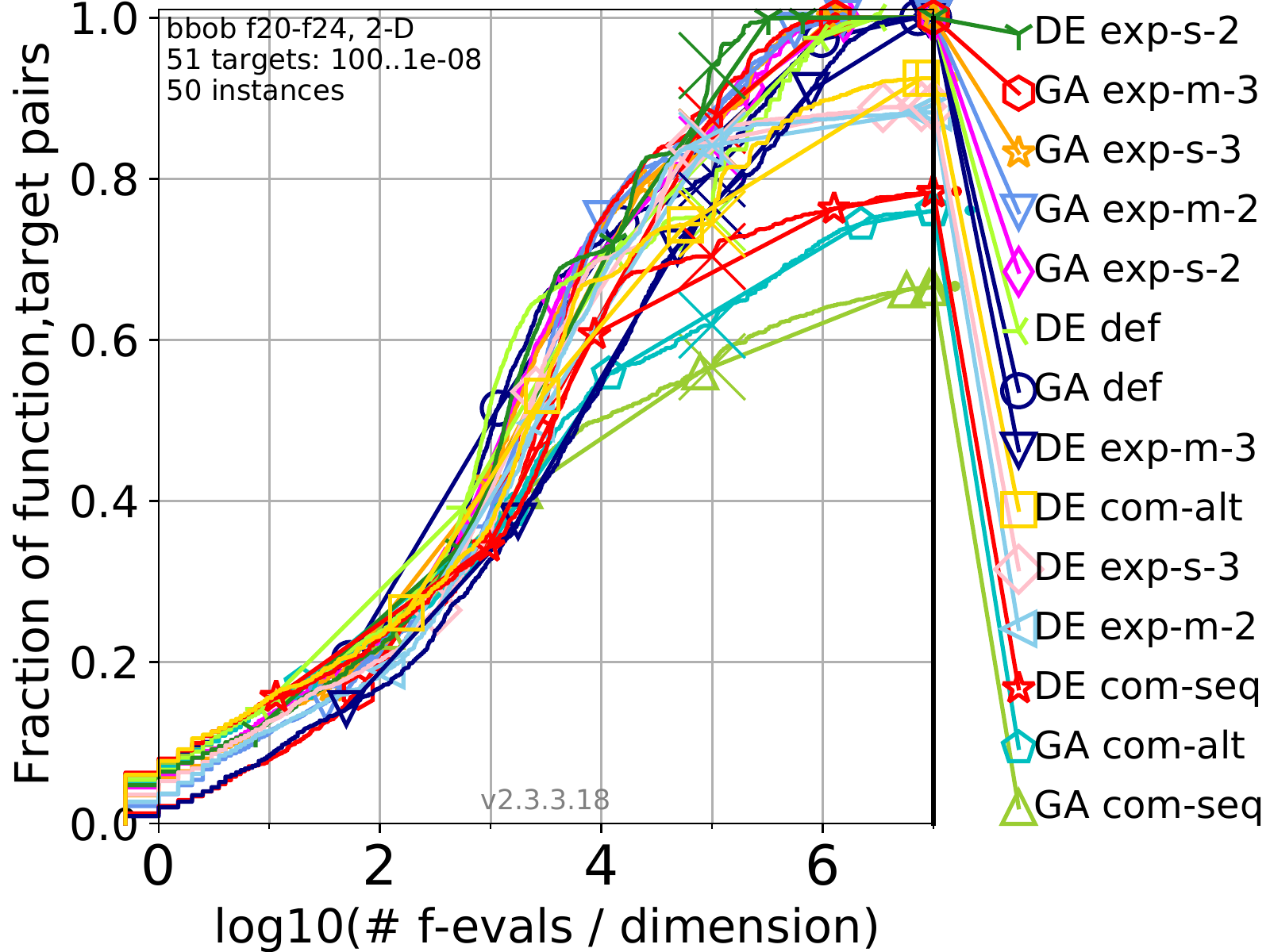}
		\caption{multi-modal functions with weak global structure}
		\label{fig:weak global structure}
	\end{subfigure}
	\caption{Empirical runtime distributions for 2 dimensional problems.}
	\label{fig:cocographs}
\end{figure*}

Figure~\ref{fig:cocographs} displays the empirical runtime distributions of the results obtained by GA and DE for the 2-dimensional function set. The x-axis shows the number of function evaluations, while the y-axis represents the proportion of problems for which the algorithm achieved the desired target values $\delta_i$, where $\delta_i \in \{10^{2}, 10^{1.8}, 10^{1.6}, 10^{1.4},\ldots, 10^{-8}\}$. More precisely, the figure depicts the ratio of the number of targets hit across all functions and the number of function, target pairs for a specific number of function evaluations.
Figure~\ref{fig:all}, presenting the total results on all functions, shows that the expansion-based approaches reach the desired function value in more cases than the default setup. For the first group of functions denoted in Figure~\ref{fig:separable}, there is little difference between the expanded and the default approach for the GA. Similar results can also be observed in other function groups, with slightly larger differences being observed only on the last group of functions.  
DE is constantly performing worse than GA, except for the multi-modal group of functions with adequate local structure, on which it obtained the desired results more often than the GA.

In all results for DE, we observe that the expansion approaches perform better than the default DE approach. This supports the observation from the GA, where the same behavior is noticed. As such, the choice of the algorithm only slightly influences the performance of the considered approaches.

Additionally, tests are performed on the two-dimensional set of functions to test the number of variables to which the original representation should be expanded. The tested expansions are $2t$, $3t$, $5t$, and $10t$. In all cases, the obtained results are quite similar, and the approaches did not show a significant deterioration in the results as the expanded number of variables increased. For that reason, we decided to keep the number of expanded variables minimal, as there is no benefit of using larger values. 

\subsection{Modeling Physical Unclonable Functions (PUFs).}

Physical Unclonable Functions (PUFs) are lightweight hardware devices commonly used in authentication schemes and anti-coun\-ter\-feit\-ing applications. PUFs use inherent manufacturing differences within every physical object to give each physical instance (a PUF) a unique identity. 
PUFs are usually divided into two categories: weak PUFs and strong PUFs. 
A strong PUF can be queried with an exponential number of challenges to receive an exponential number of responses (challenge-response pairs - CRP). Existing strong PUFs can be simulated in software, and the required parameters for such a software model can be approximated by using machine learning or evolutionary algorithms~\cite{10.1007/978-3-662-48324-4_27, 10.1145/3067695.3082535}.
Usually, strong PUFs rely on delayed-based Arbiter PUFs (APUFs) as their main building blocks for PUF constructs and protocols~\cite{7450665}. Such APUFs can be modeled by a linear function, which is at the foundation of various AI-based attacks using challenge-response pairs~\cite{DBLP:journals/tifs/Delvaux19}. 

Arbiter PUF consists of one or more chains of two 2-bit multiplexers that have identical layouts.
Each multiplexer pair is denoted as a stage, with $n$ stages in a single chain.
A single input signal is introduced to the first stage to both the bottom and top multiplexer in the pair.
The chain is fed a control signal of $n$ bits called a challenge, where each bit determines whether the two input signals in that stage would be switched (crossed over) or not.
In ideal conditions, the input signal would propagate at the same speed through each stage, and both the lower and upper signal would arrive at the arbiter (at the end of the chain) at the same time.
Due to the manufacturing inconsistencies, each delay of a multiplexer is slightly different, and the top and bottom input signals are not synchronized.
The arbiter at the end of the chain determines which signal arrived earlier and thus forms the response (0 or 1).
The response of a PUF is determined by the delay difference between the top and bottom input signal, which is, in turn, the sum of delay differences of the individual stages.
To efficiently model a PUF, one tries to determine the delay vector $w=(w_1,\ldots,w_{n+1})$ which models the delay differences in each stage. 
Lim~\cite{1561249} proposed a linear additive model that captures the APUF behavior where we require the map $f(c) = \phi$ of the applied challenge $c$ of length $n$ to a feature vector $\phi$ of length $n + 1$. The product of the feature vector and delay vector decides what signal came first, and based on it, what is the response bit $r$:
\begin{eqnarray}
\phi_i = \prod_{l=i}^{k}(-1)^{c_l}, \text{for } 1 \leq i \leq k.\\ 
%\Delta D = \vec{w}\vec{\phi}^T.\\ 
r = \begin{cases}
1 &\text{if \  $\vec{w}\vec{\phi}^T < 0$}\\
0 &\text{if \ $\vec{w}\vec{\phi}^T > 0$} 
\end{cases}
\end{eqnarray}

The optimization algorithm aims to find a delay vector that reproduces the target PUF behavior, with its actual delay values being unknown.
The delay vector is a sequence of floating-point values with $n$ elements, which correspond to $n$ stages in a PUF.
As the optimized delay vector approaches the actual one, the clone PUF model will reproduce the target PUF responses more accurately, which is the goal of the attacker.
The performance measure of the PUF model is commonly defined as the number of wrong responses in a given set of challenge-response pairs.
This value is minimized, and the lower the value, the more accurate the PUF model.

\begin{figure}
	\centering
	\includegraphics[width=0.75\textwidth]{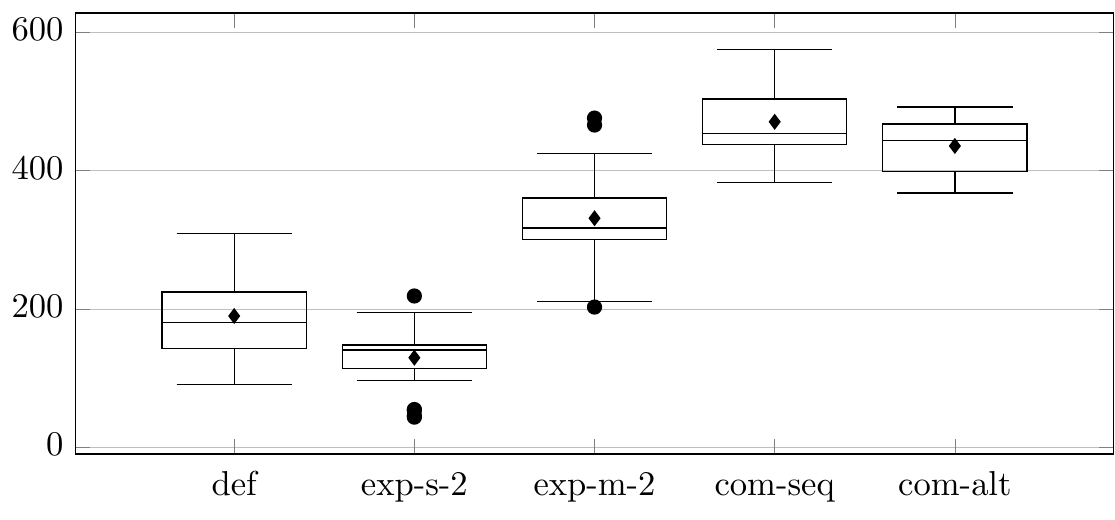}
	\caption{Results for DE optimization of PUF with chain size of 64.}
	\label{fig:puf64}
\end{figure}

Our experiments modeled PUF targets with two chain sizes, 32 and 64 elements (corresponding to 32 and 64 variables).
In Table~\ref{tab:puf}, we give the results optimizing both PUF targets with 2\,000 challenge-response pairs using GA and DE.
Although there is no significant difference for the GA, the expansion summation method slightly improved upon the default encoding.
The multiplication approach obtained slightly worse results, following the observation that the summation variant scales better with increased dimensionality, which is considerably larger in this case. We depict the results for PUF with 64 stages in Figure~\ref{fig:puf64}, where the y axis denotes the number of incorrectly modeled challenge-response pairs. Observe how the expansion procedure with summation improves slightly over the default approach, while the compression approaches work significantly worse on average. Still, the best solutions for the sequential strategy compression perform similarly to the default strategy.

\begin{table}[]
	\centering
	\small
	\caption{Results for PUFs with GA and DE for two different PUF sizes. The number of challenge response pairs is set to 2\,000.}
	\adjustbox{max width=\textwidth}{
		\begin{tabular}{@{}clllll@{}}
			\toprule
			\multicolumn{1}{l}{} & \multicolumn{1}{c}{\textbf{def}} & \multicolumn{1}{c}{\textbf{exp-s-2}} & \multicolumn{1}{c}{\textbf{exp-m-2}} & \multicolumn{1}{c}{\textbf{com-seq}} & \multicolumn{1}{c}{\textbf{com-alt}} \\ \midrule
			\textbf{GA\_2000\_32} & 38.0±15.35 & 35.0±17.72 = & 46.5±18.79 = & 73.5±68.58 - & 70.5±83.02 - \\
			\textbf{GA\_2000\_64} & 148.0±41.13 & 144.5±31.07 = & 225.5±54.25 - & 261.5±72.17 - & 299.5±83.96 - \\\midrule
			\textbf{DE\_2000\_32} & 43.0±47.19 & \cellcolor[HTML]{dfdfdf}18.0±33.73 + & 252.0±62.32 - & 440.5±56.34 - & 446.5±69.91 - \\
			\textbf{DE\_2000\_64} & 183.0±57.68 & \cellcolor[HTML]{dfdfdf}141.0±40.35 + & 317.5±63.47 - & 455.0±46.24 - & 449.5±40.05 - \\
			\bottomrule
	\end{tabular}}
	\label{tab:puf}
\end{table}

\begin{table*}[!ht]
	\small
	\caption{Results for the neural network weight optimization.}
	\label{tab:nnres}
	\adjustbox{max width=\textwidth}{
		\begin{tabular}{@{}cclllllll@{}}
			\toprule
			\multicolumn{1}{l}{$f$} &\multicolumn{1}{c}{arch} & \multicolumn{1}{c}{\textbf{def}} & \multicolumn{1}{c}{\textbf{exp-s-2}} & \multicolumn{1}{c}{\textbf{exp-s-3}} & \multicolumn{1}{c}{\textbf{exp-m-2}} & \multicolumn{1}{c}{\textbf{exp-m-3}} & \multicolumn{1}{c}{\textbf{com-seq}} & \multicolumn{1}{c}{\textbf{com-alt}} \\ \midrule
			$f_1$ & 1-5-3-1 & 1053±58.27 & \cellcolor[HTML]{dfdfdf}295.8±78.33 + & \cellcolor[HTML]{dfdfdf}271.1±71.62 + & \cellcolor[HTML]{dfdfdf}295.1±58.92 + & \cellcolor[HTML]{dfdfdf}301.3±273.0 + & 1241±208.9 - & 1184±271.5 - \\
			$f_1$ & 1-5-5-1 & 452.2±67.7 & \cellcolor[HTML]{dfdfdf}279.4±86.07 + & \cellcolor[HTML]{dfdfdf}263.1±58.31 + & \cellcolor[HTML]{dfdfdf}269.4±77.11 + & \cellcolor[HTML]{dfdfdf}286.5±91.56 + & 478.2±81.21 = & 449.2±83.05 = \\
			$f_1$ & 1-7-5-1 & 345.9±64.88 & \cellcolor[HTML]{dfdfdf}238.6±48.31 + & \cellcolor[HTML]{dfdfdf}233.6±47.45 + & \cellcolor[HTML]{dfdfdf}217.7±72.52 + & \cellcolor[HTML]{dfdfdf}231.4±63.95 + & 366.5±57.02 = & 340.2±50.27 = \\
			$f_1$ & 1-7-7-1 & 324.4±60.28 & \cellcolor[HTML]{dfdfdf}235.4±41.78 + & \cellcolor[HTML]{dfdfdf}201.9±42.75 + & \cellcolor[HTML]{dfdfdf}233.5±66.67 + & \cellcolor[HTML]{dfdfdf}241.1±72.04 + & 348.6±65.3 - & 378.6±75.53 - \\
			$f_2$ &2-5-3-1 & 18.45±3.883 & 19.62±3.813 = & 19.54±4.06 = & 23.76±3.964 - & 20.22±5.248 - & 22.59±2.327 - & 24.47±3.195 - \\
			$f_2$ &2-7-5-1 & 14.37±2.615 & 14.67±3.53 = & 14.31±3.712 = & 14.89±4.679 = & 13.73±6.559 = & 18.77±3.324 - & 18.15±3.842 - \\
			$f_2$ & 2-7-7-1 & 14.31±3.172 & 13.13±4.356 = & 14.41±3.779 = & 14.3±4.345 = & 11.78±5.178 = & 18.0±2.73 - & 18.1±3.533 - \\
			$f_2$ &2-5-5-1 & 17.81±4.17 & 17.61±4.347 = & 18.92±3.614 = & 18.4±4.71 = & 19.81±6.001 = & 22.47±3.941 - & 21.75±3.747 - \\
			$f_3$ &1-2-2-1 & 5.604±0.574 & 5.614±1.246 = & \cellcolor[HTML]{dfdfdf}5.53±0.616 + & 5.728±5.562 = & 5.82±0.994 - & 5.952±6.089 - & 6.018±8.752 - \\
			$f_3$ &1-3-3-1 & 5.525±0.572 & \cellcolor[HTML]{dfdfdf}4.691±0.826 + & \cellcolor[HTML]{dfdfdf}5.315±0.645 + & 5.37±1.884 = & 5.439±1.44 = & 5.508±6.551 = & 5.57±6.626 = \\
			$f_3$ &1-5-5-1 & 4.263±0.961 & 4.166±1.04 = & \cellcolor[HTML]{dfdfdf}3.803±0.922 + & 4.287±0.906 = & 4.122±0.822 = & 4.168±1.695 = & 4.215±7.981 = \\
			$f_3$ &1-7-5-1 & 3.761±0.682 & \cellcolor[HTML]{dfdfdf}3.427±0.877 + & 3.609±0.747 = & 3.797±1.006 = & 3.619±1.19 = & 4.611±1.275 - & 4.022±10.11 = \\ \bottomrule
	\end{tabular}}
\end{table*}

\subsection{Optimizing Neural Network Weights.}
Artificial neural networks (ANNs) are a widely used model applied for various problem types. However, determining the multiplexer's optimal weights of a neural network for a given problem is a difficult optimization problem. As a result, various algorithms are used to optimize the weights of ANNs, ranging from gradient-based to evolutionary methods. We apply the proposed approaches for optimizing the weights of ANNs for three selected regression problems. The considered problems are:
\begin{align}
&f_1(x)=3\cdot \sin(x)+x. \\
&f_2(x,y)=x+y. \\
&f_3(x)=x\cdot \sin(x).
\end{align}
We selected these problems as they include both linear and nonlinear functions. For each problem, the number of training samples is between 250 and 300. We consider a simple feed-forward fully-connected ANN consisting of two hidden layers. The notation \textit{a-b-c-d} is used to denote the architecture of a network with \textit{a} nodes in the input layer, \textit{b} and \textit{c} nodes in the first and second hidden layers, and \textit{d} nodes in the output layer.
The Sigmoid function is used as the activation function in the hidden layers, whereas the linear sum function is used in the output layer. The following architectures are applied for the first and second regression problems: 1-5-3-1, 1-5-5-1, 1-7-5-1, 1-7-7-1. For the second function, the input layer consisted of two nodes instead of one. Since the third regression problem was less difficult to solve, it is used to test some smaller architectures: 1-2-2-1, 1-3-3-1, 1-5-5-1, 1-7-5-1. Each experiment is run 30 times to obtain statistically significant results. The domain is set to $[-5,5]$, and the termination criterion is set to 100\,000 function evaluations. The fitness function of this problem is defined as the mean squared error between the real values and the ANN output. Note that the problems and architectures of the considered neural networks are simple by today's deep learning standards, but they were selected to test the feasibility of the proposed approach for such problems.

\begin{figure}[!ht]
	\centering
	\begin{subfigure}[t]{0.75\textwidth}
		\centering
		\includegraphics[width=\textwidth]{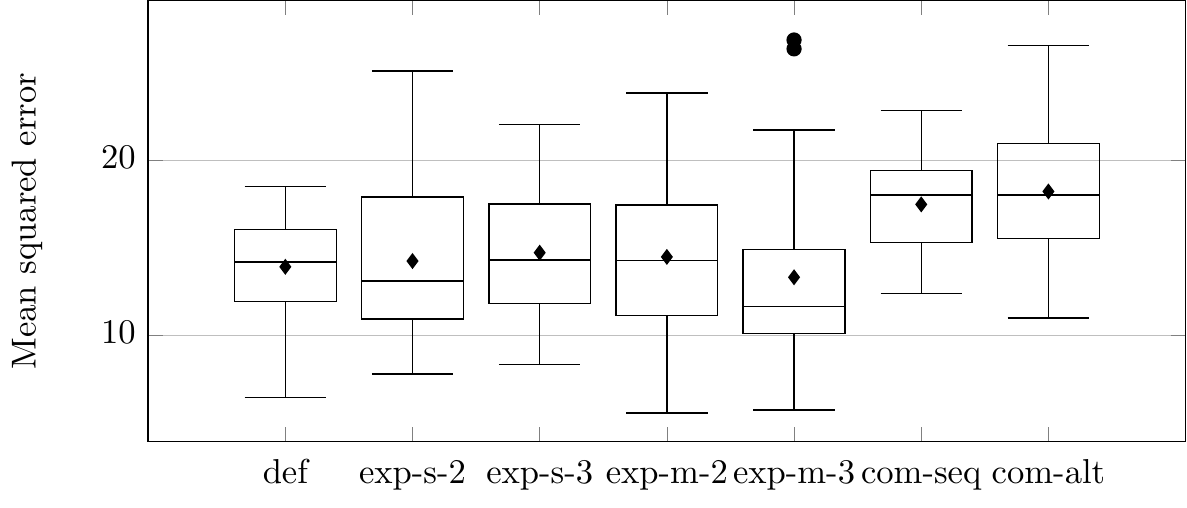}
		\caption{Optimizing function $x+y$ with the 2-7-7-1 architecture.}
		\label{fig:nnbox1}
	\end{subfigure}
	\begin{subfigure}[t]{0.75\textwidth}
		\centering
		\includegraphics[width=\textwidth]{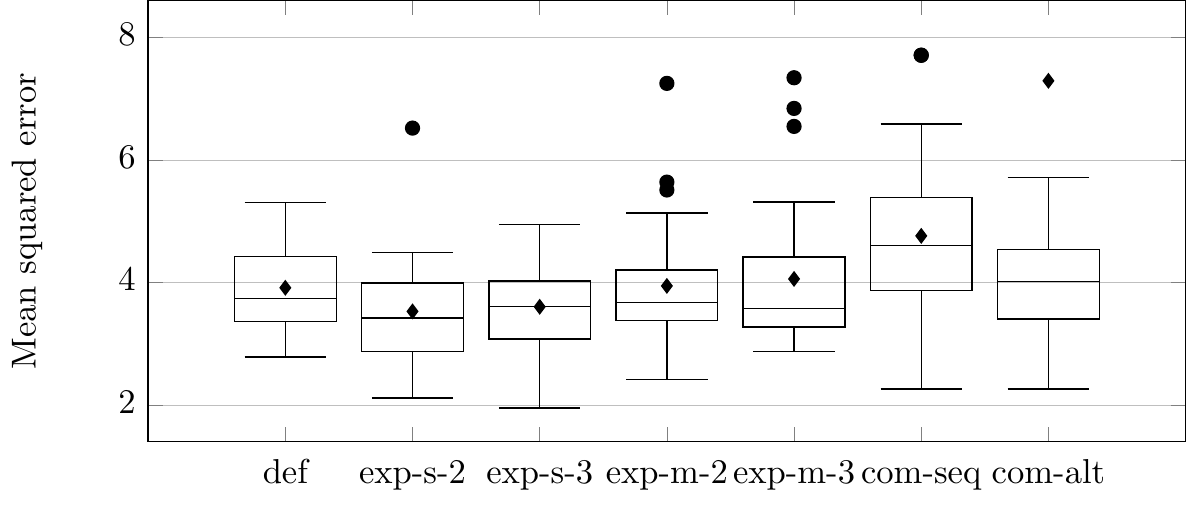}
		\caption{Optimizing function $xsin(x)$ with the 1-7-7-1 architecture.}
		\label{fig:nnbox2}
	\end{subfigure}
	\begin{subfigure}[t]{0.75\textwidth}
		\centering
		\includegraphics[width=\textwidth]{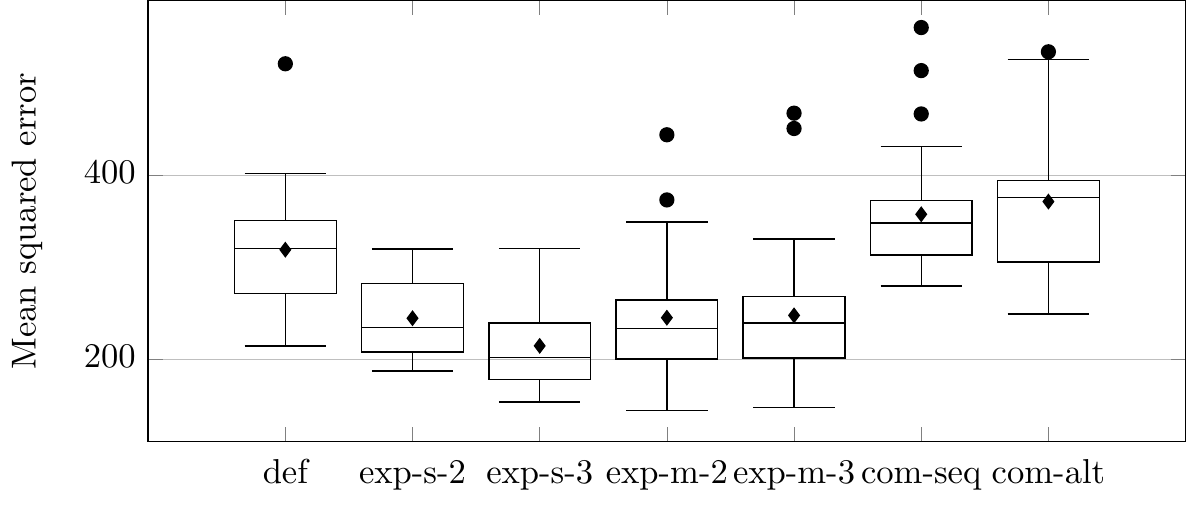}
		\caption{Optimizing function $3sin(x)+x$ with the 1-7-7-1 architecture.}
		\label{fig:nnbox3}
	\end{subfigure}
	\caption{Distribution of the results obtained for optimizing neural network weights with GA.}
	\label{fig:nn_results}
\end{figure}

Table~\ref{tab:nnres} displays the results obtained by GA for each regression problem and the selected architectures, while Figure~\ref{fig:nn_results} represents the distribution of solutions. Each row contains the median MSE value of the 30 executions and the standard deviation, denoted after the $\pm$ sign. The results support what was already demonstrated in previous experiments: the compression-based approaches again obtain inferior performance compared to all other tested approaches. On the other hand, the summation-based expansion approach achieved equally good or significantly better results than the default approach in all experiments. The multiplication-based approaches also performed well, although slightly worse than the summation-based ones. Surprisingly, for this problem, the increase of the number of variables, i.e., weights, did not affect the expansion approaches' performance, unlike in the case of the COCO benchmarks. Therefore, the results also demonstrate that the expansion approaches' performance depends on the considered optimization problem.

The probable reason why the compression-based methods achieve inferior results is that two variables are fused into one. This means that changes are being performed on both original variables at the same time as the changes are performed on the compressed variable. As such, it is quite difficult to ensure that both variables are being changed in a meaningful way, which would lead us closer to the minimum, or that only a single of those two variables is updated. The compression-based methods also suffer from a reduced precision since the same space is used to store more than one variable at a time. This can limit the algorithms since they cannot perform an equally fine-grained search as without the genotype reduction. On the other hand, the expansion approaches seem to perform better due to a larger degree of freedom. 
However, in the expansion approaches, due to each variable being expressed with multiple values, the algorithm now has an infinite number of combinations in which it can represent a single value with several ones. Although the search space also increases, it seems that this additional freedom proves to be beneficial and allows the algorithms to find new paths towards the minimum.

\section{Conclusions and Future Work}
\label{sec:conclusions}

This paper discusses how to expand or compress genotypes for continuous optimization and EAs. We compare four evolutionary algorithms' performance, four strategies for reconstructing the original genotype after compression/expansion, and three sets of problems.
Our analysis shows that compression works poorly in all the tested cases, while expansion manages to outperform the default encoding in numerous settings. We find the summation-based expansion strategy to be especially promising.

As the approach we discuss here is novel, there are multiple research directions one could follow. While we discuss the continuous optimization problems where we see the improvements stemming from the genotype expansions, it is not difficult to imagine problems where such an approach would not be beneficial. We plan to investigate such problems and understand what the differences are.
Also, we considered a scenario where every variable is either expanded or compressed. It would be interesting to see what happens when only a subset of variables is adapted in this way. This could provide a trade-off between the improvements in the performance and the size of the genotype.

\bibliographystyle{abbrv}
\bibliography{bibliography}

\end{document}